\documentclass[runningheads]{llncs}
\usepackage{graphicx}
\usepackage{algorithmic}
\usepackage[ruled,vlined]{algorithm2e}
\usepackage{xcolor}
\begin{document}
\title{Pyrus Base: An Open Source Python Framework for the RoboCup 2D Soccer Simulation}

\author{
Nader Zare\inst{1}\inst{*}\and 
Aref Sayareh\inst{3}\inst{*}\and
Omid Amini\inst{1}\and
Mahtab Sarvmaili\inst{1}\and
Arad Firouzkouhi\inst{4}\and
Stan Matwin\inst{1,}\inst{2} \and
Amilcar Soares\inst{3}}
\authorrunning{N. Zare, A.Sayareh, et al.}

\institute{
Institute for Big Data Analytics, Dalhousie University, Halifax, Canada\\
\and
Institute for Computer Science, Polish Academy of Sciences, Warsaw, Poland\\
\and
Memorial University of Newfoundland, St. John's, Canada\\
\and
Amirkabir University of Technology, Iran\\
\email{nader@cyrus2d.com},\\
\email{\{asayareh, amilcarsj\}@mun.ca},\\
\email{\{omid.amini, mahtab.sarvmaili\}@dal.ca},\\ 
\email{arad.firouzkouhi@aut.ac.ir},\\ 
\email{stan@cs.dal.ca} 
}

\def\thefootnote{*}\footnotetext{These authors contributed equally to this work.}

% \def\thefootnote{*}\footnotetext{These authors contributed equally to this work.}

% \addauthor{Name\thefootnote{}}{email/homepage}{INSTITUTION_CODE}

%
\maketitle              % typeset the header of the contribution
\begin{abstract}

Soccer, also known as football in some parts of the world, involves two teams of eleven players whose objective is to score more goals than the opposing team.
To simulate this game and attract scientists from all over the world to conduct research and participate in an annual computer-based soccer world cup, Soccer Simulation 2D (SS2D) was one of the leagues initiated in the RoboCup competition.
In every SS2D game, two teams of 11 players and one coach connect to the RoboCup Soccer Simulation Server and compete against each other.
Over the past few years, several C++ base codes have been employed to control agents' behavior and their communication with the server. 
Although C++ base codes have laid the foundation for the SS2D, developing them requires an advanced level of C++ programming. 
C++ language complexity is a limiting disadvantage of C++ base codes for all users, especially for beginners. 
To conquer the challenges of C++ base codes and provide a powerful baseline for developing machine learning concepts, we introduce Pyrus, the first Python base code for SS2D.
Pyrus is developed to encourage researchers to efficiently develop their ideas and integrate machine learning algorithms into their teams. Pyrus base is open-source code, and it is publicly available under MIT License on GitHub\footnote{https://github.com/Cyrus2D/Pyrus2D}.


\end{abstract}

\section{Introduction}
Soccer is one of the most popular team sports in the world. The main purpose of this game is to achieve more goals than the opposing team in a multi-player, real-time, strategic, and partially observable game. 
Players must manage different tactical and technical strategies in addition to cooperative behavior \cite{soccer1,soccer2,soccer3}.
Considering the challenging and exciting nature of soccer, computer simulation of this game creates an interesting environment for developing machine learning algorithms that can address problems such as multi-agent learning \cite{deepmind}. 
On this matter, the World Cup Robot Soccer Initiative was established to create a realistic environment similar to the real game of soccer, encouraging researchers to employ robotics and artificial intelligence (AI) to solve a wide range of problems\cite{noda1996soccer}. The primary goal of this tournament is to design a robotic team to compete against the best human team by 2050 \cite{burkhard2002}. 
It was during IJCAI-97 when the first RoboCup competition was held, and three competition tracks were offered: the real robot league, the software robot competition, and the expert robot competition\cite{robo1997,kitano1997robocup}. 
The Soccer Simulation 2D league (SS2D) presents a wide range of research challenges, including autonomous decision-making, communication, coordination, tactical planning, collective behavior, and opponents' actions prediction \cite{ref1,ref2,ref3,ref4,ref5,ref6,ref7,ref8,cyrussamposiom}.

\begin{figure}[ht]
    \centering
    \includegraphics[scale=0.450]{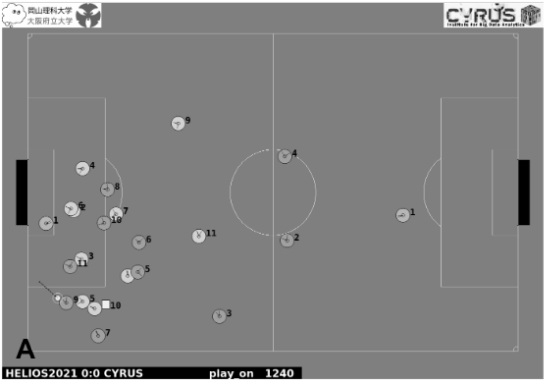} \\
    \caption{Visualization of Soccer Simulation 2D League.}
    \label{fig:2d}
\end{figure}

In this league, the RoboCup Soccer Simulation Server (RCSSServer) is responsible for executing and managing a 2D soccer game between two teams of twelve autonomous software programs(agents). Agents receive relative and noisy information about the environment, and based on their logic and algorithms, they produce basic commands (such as dashing, turning, or kicking) to influence the environment. A visual example of the game is shown in Figure \ref{fig:2d}. 

Developing an agent program from scratch in SS2D is a challenging task that requires resolving technical issues such as stable network communication, synchronization, and world modeling. However, having a base code of agents can accelerate the progress of developing effective multi-agent teamwork techniques.
The base code is responsible for handling the complexity of the game, such as communicating with the server, modeling the server world, and making multi-agent decisions. This operational base code is essential for advancing research in multi-agent systems in the context of soccer simulation\cite{agent2d,cyrus2d}.

% In this paper, we will describe released base codes and sample teams, their benefits and challenges. We will then introduce Pyrus, the first Python base code in SS2D.

A description of released base codes and sample teams, their advantages and disadvantages will be presented in this paper. 
We will then introduce Pyrus, the first Python base code in SS2D.

\section{Background}
Several teams have released base codes for RoboCup soccer simulation, including the "CMUnited" team from Carnegie Mellon University (USA)\cite{stonerobo2000,stonerobo99}, the "UvA Trilearn" team from the University of Amsterdam (The Netherlands)\cite{UTR}, the "MarliK" team from the University of Guilan (Iran)\cite{MKR}, the "HELIOS" team from AIST Information Technology Research Institute (Japan), the "Wrighteagle" team from University of Science and Technology of China\cite{wrighteagle}, "Gliders2d" team from the University of Sydney (Australia)\cite{glbase,glbase2}, and "Cyrus2d" team from the Dalhousie University (Canada)\cite{cyrussamposiom,cyrus2d}.

Using these base codes, teams have been able to focus on developing high-level strategies and algorithms rather than dealing with low-level implementation issues.
Among the published base codes, the "Helios" base code is the most referenced, and the "Cyrus2D" is the most powerful base code for SS2D players. It was implemented based on "Helios", and "Gliders2D" and significant features of the CYRUS team\footnote{The championship of RoboCup 2021}. 

All of the mentioned bases are developed in C++, a mid-level, high-speed, and efficient programming language in handling computationally demanding tasks.
These base code Object-oriented programming (OOP) structures are beneficial by making the base codes more maintainable.

% C++ has been the primary language for developing base codes for RoboCup soccer simulation, 
% w.r.t its speed and efficiency in handling computationally demanding tasks. 

In contrast, due to C++ syntax complexity (such as pointers), comprehending, debugging, and implementing new algorithms is difficult, especially for beginners. In addition, these base codes support deprecated server features, making them challenging to understand.

On the other hand, Python is a more user-friendly programming language that prioritizes readability and simplicity.
With its extensive community support, simpler debugging capabilities, and easy-to-use syntax, Python's popularity has increased significantly in recent years.
Although Python may not be as fast as C++, it offers a considerable number of libraries and frameworks that can speed up development and reduce programming time.
As a result, many developers are now exploring Python as an alternative language for creating soccer simulation agents\cite{ggsoccer}.

In this regard, Half Field Offense (HFO) framework was developed to work as a Python interface on the Helios C++ base \cite{hfo}. This framework uses the "Helios" base to connect to the server, receive and store observations, and send actions. Due to its open-source availability and Python interface, the "HFO" encourages machine learning researchers to work on the soccer simulation 2D server. However, to apply any changes to the base code, researchers must work with C++, since HFO is tied to the "Helios" base code. Additionally, "HFO" is not compatible with the latest version of the soccer simulation server.

To break any reliance on C++ base codes, we have designed the first complete Python SS2D base code called PYRUS, which has been implemented from scratch using this language. In the next section, we will explain the details of Pyrus and the corresponding geometry library a.k.a PyrusGeom library.

\section{Pyrus}
\subsection{Pyrus Geom Library}
"Pyrus Geom Library" is a Python library that simplifies two-dimensional geometric calculations in Python for the Pyrus. 
% \textcolor{red}{This library can be used in other libraries and applications as well such as teams, bases and log analyzers.}
Many geometrical objects are implemented in this library, including Angle2D, Circle2D, ConvexHull, Line2D, Matrix2D, Polygon2D, Ray2D, Rect2D, Region2D, Sector2D, Segment2D, Size2D, Triangle2D, and Vector2D\footnote{Researchers can install and use this library using Python Pip or https://github.com/Cyrus2D/PyrusGeom/}.
This library can be adopted for other Python applications and domains.

\subsection{Pyrus Base Code}
Members of the CYRUS SS2D team has commenced the Pyrus project in 2019. 
Besides supporting the latest features of the RCSSServer (version 18), the Pyrus base code includes simple offensive and defensive decision-making algorithms to choose the appropriate action such as passing, dribbling, intercepting, blocking, etc.

In addition to supporting the latest version (version 18) of the player, coach, and trainer in the server, we will keep updating the project to support future server versions. 

In this base code, players can support full state (observation without noise) and normal observation mode in the synchronous and standard timer of the server.
While the general design of the base code is similar to "Helios", structural simplifications have made it easier for beginners to adopt it for their team. For example, deprecated features of servers were not implemented in the base.
In the following subsections, we will discuss the features, challenges, structure, and several major algorithms implemented in Pyrus.

\subsection{Features}
The Pyrus base code is implemented in Python 3.9. 
Due to Python's simple coding style, researchers can focus on the development of their ideas rather than language structural complexity.
Also, debugging the implemented algorithms is more efficient than C++ implementation. 
Python is one of the most popular languages for machine learning, and it comes with a huge variety of well-implemented machine-learning libraries that researchers can use directly in Pyrus. 
Although Pyurs contains about 30k lines of code, it is easier to read and understand the framework and its internal processing w.r.t the C++ bases such as "Helios" or "Cyrus2D" (150k lines). 

In addition to supporting the Python logging framework, this code also supports the debug-client and file-logging in soccer window debugging systems.  
Using the Python multi-processing and shared memory packages, players and trainers can distribute their data. As an example, this feature is useful during the training phase of RL agents, which requires passing the experiences among the agents.
% \textcolor{cyan}{Players and trainers can share data using the Python multi-process and shared memory framework, simplifying the reinforcement learning training phase.} 
\subsection{Challenges}
Considering the performance of Python, Pyrus processing time takes longer than the C++ base codes, therefore we are planning to address this problem by compiling the code with Cython\cite{cython}. 
Cython is designed to ameliorate the execution time of a Python program to a C-like performance.
Additionally, to enhance the processing speed of mathematical calculations, researchers can employ libraries that are supported by C/C++ backends, such as NumPy\cite{numpy}.

\subsection{Structure}
\begin{figure}[ht]
    \centering
    \includegraphics[scale=0.1]{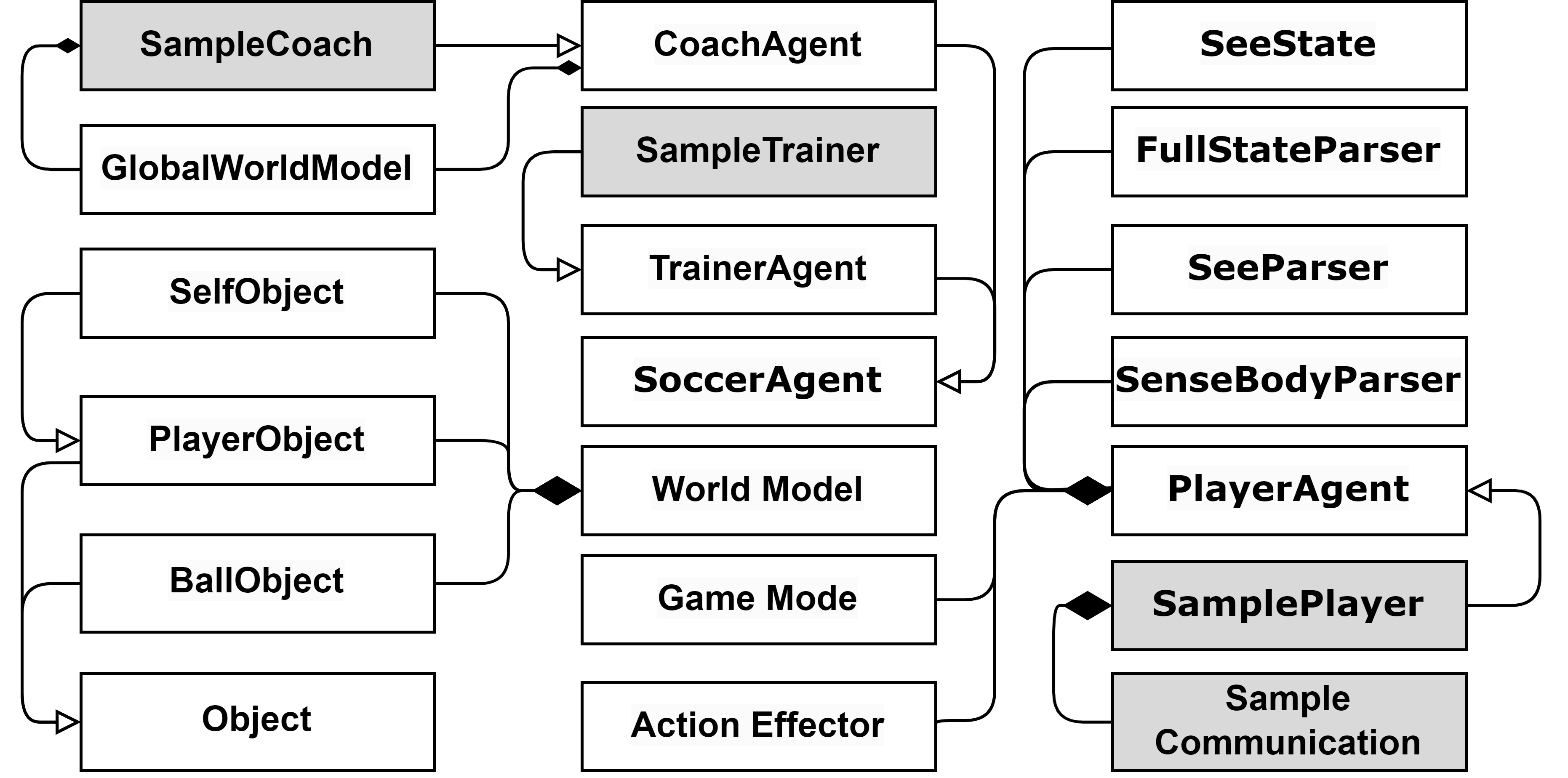} \\
    \caption{UML of main Classes in Pyrus}
    \label{fig:class}
\end{figure}

The Pyrus structure is based on Object Oriented Programming techniques and it is similar to the "Helios" base code. The structure of Pyrus is shown in Figure \ref{fig:class}.
% The details of several important classes are provided in the following.

% \subsubsection{SoccerAgent.}
% This abstract class is designed to be a foundation for other agent classes. This class provides methods for
% \begin{itemize}
%     \item managing the connection between the server and an agent.
%     \item updating the state of the agent and the field.
%     \item making decisions based on these observations.
% \end{itemize}
% Inherited classes include "PlayerAgent," "CoachAgent," and "TrainerAgent."

% \subsubsection{WorldModel.}
% The player's observation is stored and modeled using this class. The "WorldModel" class contains information about the ball and the players. 

% \subsubsection{InterceptTable.}
% The class calculates the number of necessary cycles for players to intercept the ball based on their locations and velocities by simulating their future actions. This enables precise prediction of player movements.
 
\subsection{Algorithm}
\begin{figure}[ht]
    \centering
    \includegraphics[scale=0.090]{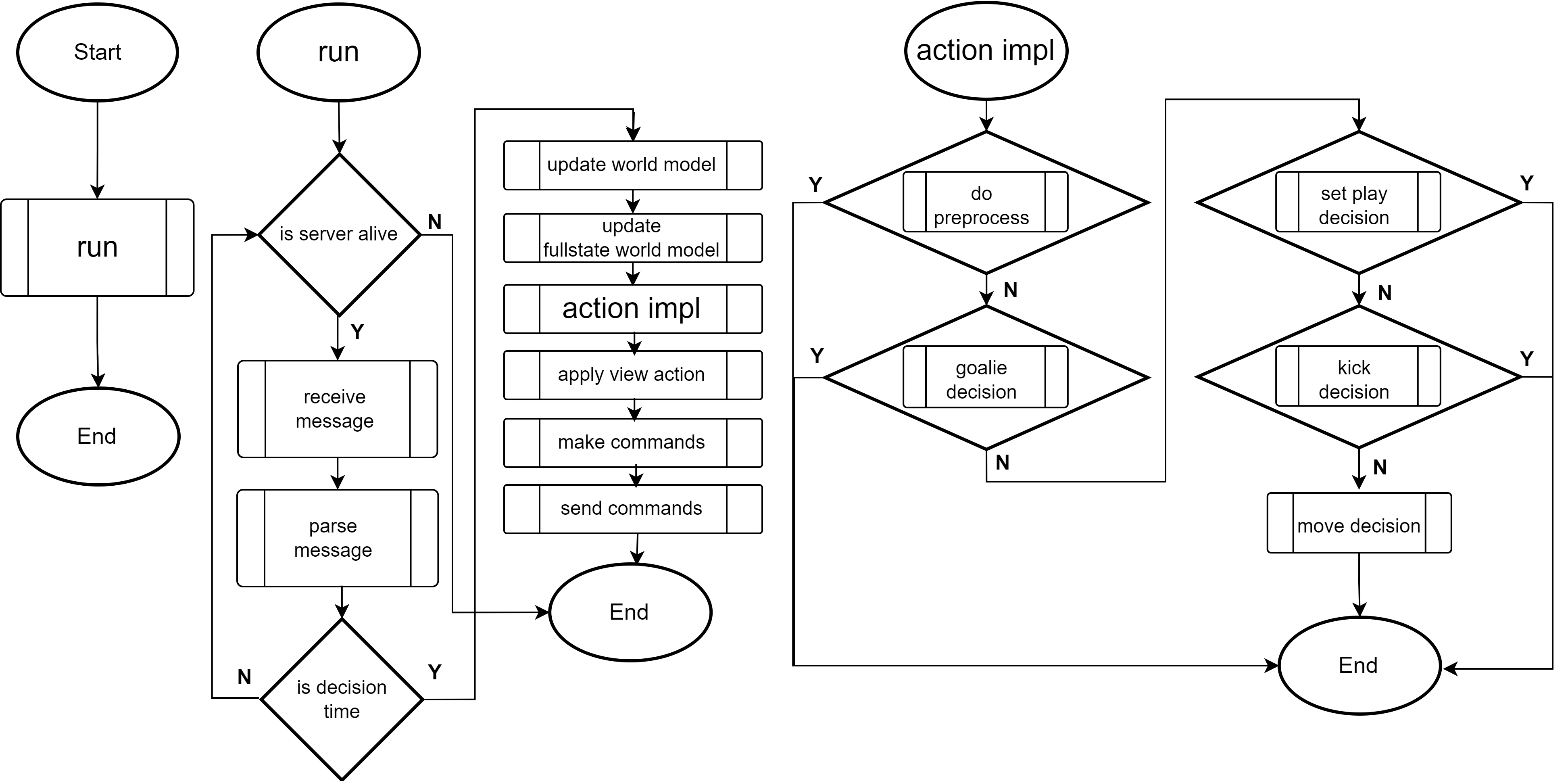} \\
    \caption{FlowChart of Pyrus}
    \label{fig:flow}
\end{figure}

In order to maintain the simplicity of the code, Pyrus does not support old or deprecated features of the server, such as asynchronous see mode.
In order to complete one cycle of the game, a Pyrus agent receives messages from the server, parses them, updates variables, makes decisions, and sends commands to the server. The simplified flowchart of the Pyrus is shown in Figure \ref{fig:flow}.

% \begin{algorithm}[!htb]
%     \SetAlgoLined
%     \emph{\textbf{function action\_impl}():} \\
%         \quad \emph{\textbf{if }}{do\_pre\_process}():\\
%             \quad \quad \emph{{return}}\\
%         \quad \emph{\textbf{if }}{goalie\_decision}():\\
%             \quad \quad \emph{{return}}\\
%         \quad \emph{\textbf{if }}{setplay\_decision}():\\
%             \quad \quad \emph{{return}}\\
%         \quad \emph{\textbf{if }}{kick\_decision}():\\
%             \quad \quad \emph{{return}}\\
%         \quad \emph{\textbf{}}{move\_decision}():\\
%     \quad\\
%     \emph{\textbf{function run}():} \\
%         \quad \emph{\textbf{while }}{server \textbf{is} alive}:\\
%             \quad \quad \emph{\textbf{}}receive\_message()\\
%             \quad \quad \emph{\textbf{}}parse\_message()\\
%         \quad \quad \emph{\textbf{if }}{is\_decision\_time()}:\\
%             \quad \quad \quad \emph{{update\_world\_model()}}\\
%             \quad \quad \quad \emph{{update\_full\_state\_world\_model()}}\\
%             \quad \quad \quad \emph{{action\_impl()}}\\
%             \quad \quad \quad \emph{{apply\_view\_action()}}\\
%             \quad \quad \quad \emph{{apply\_neck action()}}\\
%             \quad \quad \quad \emph{{make\_commands()}}\\
%             \quad \quad \quad \emph{{send\_commands()}}\\
            
%     \caption{Land Attended Rammer-Douglas-Peucker}
%     \label{LAR}
% \end{algorithm}

\section{Conclusion and Future Works}

% Using these base codes, teams have been able to focus on developing high-level strategies and algorithms rather than dealing with low-level implementation issues.
The Robocup has provided a platform for researchers to employ robotics and AI solutions to solve various problems. The Soccer Simulation 2D league presents a variety of research challenges that require multi-agent decision-making, communication, coordination, tactical planning, and behavior prediction of the opponents. 
In order to decrease the amount of time spent dealing with low-level implementation issues in working with the RCSSServer, several C++ base codes have been released. However, C++ is not the most suitable language for implementing deep learning and reinforcement learning algorithms, and Python has become the most popular programming language for machine learning research. 
In this paper, we presented the Pyrus base code, the first Python sample team for the SS2D League. This base code has several advantages: straightforward syntax, extensible structure and fast debugging time, and availability of various machine-learning libraries. By utilizing Pyrus, more researchers can use RoboCup Soccer Simulation as a stable environment for machine learning researches.

% \section{Future works}
as a part of our future plan, we aim to enhance localization and decision-making algorithms to help beginners. To facilitate the usage of our framework, we intend to create a package to use Pyrus agents and RoboCup Soccer Simulation Server as an environment in the OpenAI gym toolkit\cite{gym}. Additionally, we plan to implement a Python monitor and log analyzer software to improve the usability of Pyrus.


\begin{thebibliography}{8}

\bibitem{soccer1}
Bangsbo J, Peitersen B. Soccer systems and strategies. Human Kinetics; 2000.

\bibitem{soccer2}
Pollard R, Reep C. Measuring the effectiveness of playing strategies at soccer. Journal of the Royal Statistical Society: Series D (The Statistician). 1997 Dec;46(4):541-50.

\bibitem{soccer3}
Rein R, Memmert D. Big data and tactical analysis in elite soccer: future challenges and opportunities for sports science. SpringerPlus. 2016 Dec;5(1):1-3.

\bibitem{burkhard2002}
Burkhard, H.D., Duhaut, D., Fujita, M., Lima, P., Murphy, R., Rojas, R.: The road to RoboCup 2050. IEEE Robotics Automation Magazine 9(2), 31–38 (Jun 2002)

\bibitem{deepmind}
Liu, S., Lever, G., Wang, Z., Merel, J., Eslami, S.M., Hennes, D., Czarnecki, W.M., Tassa, Y., Omidshafiei, S., Abdolmaleki, A. and Siegel, N.Y., 2021. From motor control to team play in simulated humanoid football. arXiv preprint arXiv:2105.12196.

\bibitem{noda1996soccer}
Noda, I. and Matsubara, H., 1996, November. Soccer server and researches on multi-agent systems. In Proceedings of the IROS-96 Workshop on RoboCup (pp. 1-7).

\bibitem{robo1997}
Kitano, H., Asada, M., Kuniyoshi, Y., Noda, I. and Osawa, E., 1997, February. Robocup: The robot world cup initiative. In Proceedings of the first international conference on Autonomous agents (pp. 340-347).

\bibitem{kitano1997robocup}
Kitano, H., Asada, M., Kuniyoshi, Y., Noda, I., Osawa, E. and Matsubara, H., 1997. RoboCup: A challenge problem for AI. AI magazine, 18(1), pp.73-73.

\bibitem{ref1}
Noda, I., Stone, P.: The RoboCup Soccer Server and CMUnited Clients: Implemented Infrastructure for MAS Research. Autonomous Agents and Multi-Agent Systems 7(1–2), 101–120 (July–September 2003)

\bibitem{ref2}
Riley, P., Stone, P., Veloso, M.: Layered disclosure: Revealing agents’ internals. In: Castelfranchi, C., Lesperance, Y. (eds.) Intelligent Agents VII. Agent Theories, Architectures, and Languages — 7th. International Workshop, ATAL-2000, Boston, MA, USA, July 7–9, 2000, Proceedings. Lecture Notes in Artificial Intelligence, Springer, Berlin, Berlin (2001)

\bibitem{ref3}
Stone, P., Riley, P., Veloso, M.: Defining and using ideal teammate and opponent models. In: Proc. of the 12th Annual Conf. on Innovative Applications of Artificial Intelligence (2000)

\bibitem{ref4}
Butler, M., Prokopenko, M., Howard, T.: Flexible synchronisation within RoboCup environment: A comparative analysis. In: RoboCup 2000: Robot Soccer World Cup IV. pp. 119–128. Springer, London, UK (2001)

\bibitem{ref5}
Reis, L.P., Lau, N., Oliveira, E.: Situation based strategic positioning for coordinating a team of homogeneous agents. In: Balancing Reactivity and Social Deliberation in Multi-Agent Systems, From RoboCup to Real-World Applications. pp. 175–197. Springer (2001)

\bibitem{ref6}
Prokopenko, M., Wang, P.: Disruptive Innovations in RoboCup 2D Soccer Simulation League: From Cyberoos’98 to Gliders2016. In: Behnke, S., Sheh, R., Sariel, S., Lee, D.D. (eds.) RoboCup 2016: Robot World Cup XX [Leipzig, Germany, June 30 - July 4, 2016]. Lecture Notes in Computer Science, vol. 9776, pp. 529–541. Springer (2017)

\bibitem{ref7}
Prokopenko, M., Wang, P., Marian, S., Bai, A., Li, X., Chen, X.: Robocup 2d soccer simulation league: Evaluation challenges. In: Akiyama, H., Obst, O., Sammut, C., Tonidandel, F. (eds.) RoboCup 2017: Robot World Cup XXI [Nagoya, Japan, July 27-31, 2017]. Lecture Notes in Computer Science, vol. 11175, pp. 325–337. Springer (2018) 

\bibitem{ref8}
Prokopenko, M., Wang, P., Obst, O.: Gliders2015: Opponent avoidance with bio-inspired flocking behaviour. In: RoboCup 2015 Symposium and Competitions: Team Description Papers, Hefei, China, July 2015 (2015)

\bibitem{cyrussamposiom}
Zare, N., Sayareh, A., Sarvmaili, M., Amini, O., Matwin, S., Soares, A.: Engineering Features to Improve Pass Prediction in 2D Soccer Simulation Games. In: RoboCup 2021: Robot World Cup XXIV, Springer (2021)

\bibitem{agent2d}
Akiyama, H., Nakashima, T.: Helios base: An open source package for the robocup soccer 2d simulation. In Robot Soccer World Cup 2013 Jun 24 (pp. 528-535). Springer, Berlin, Heidelberg.

\bibitem{cyrus2d}
Zare, N., Amini, O., Sayareh, A., Sarvmaili, M., Firouzkouhi, A., Rad, S.R., Matwin, S. and Soares, A., 2023. Cyrus2D Base: Source Code Base for RoboCup 2D Soccer Simulation League. In RoboCup 2022: Robot World Cup XXV (pp. 140-151). Cham: Springer International Publishing.

\bibitem{stonerobo2000}
Stone, P., Asada, M., Balch, T.R., Fujita, M., Kraetzschmar, G.K., Lund, H.H., Scerri, P., Tadokoro, S., Wyeth, G.: Overview of robocup-2000. In: Stone, P., Balch, T.R., Kraetzschmar, G.K. (eds.) RoboCup 2000: Robot Soccer World Cup IV. Lecture Notes in Computer Science, vol. 2019, pp. 1–28. Springer (2000)

\bibitem{stonerobo99}
Stone, P., Riley, P., Veloso, M.: The CMUnited-99 champion simulator team. In: Veloso, M., Pagello, E., Kitano, H. (eds.) RoboCup-99: Robot Soccer World Cup III, Lecture Notes in Artificial Intelligence, vol. 1856, pp. 35–48. Springer Verlag, Berlin (2000)

\bibitem{UTR}
Kok, J.R., Vlassis, N., Groen, F.: UvA Trilearn 2003 team description. In: Polani, D., Browning, B., Bonarini, A., Yoshida, K. (eds.) Proceedings CD RoboCup 2003. Springer, Padua (2003)

\bibitem{wrighteagle}
Bai A, Chen X, MacAlpine P, Urieli D, Barrett S, Stone P. Wrighteagle and ut austin villa: RoboCup 2011 simulation league champions. InRoboCup 2011: Robot Soccer World Cup XV 15 2012 (pp. 1-12). Springer Berlin Heidelberg.

\bibitem{glbase}
Prokopenko, M., Wang, P.: Gliders2d: Source Code Base for RoboCup 2D Soccer Simulation League. CoRR abs/1812.10202 (2018)

\bibitem{glbase2}
Prokopenko, M. and Wang, P., 2019, July. Fractals2019: Combinatorial optimisation with dynamic constraint annealing. In Robot World Cup (pp. 616-630). Springer, Cham.

\bibitem{MKR}
Tavafi, A., Nozari, N., Vatani, R., Yousefi, M.R., Rahmatinia, S., Pirdir, P.: MarliK 2012 Soccer 2D Simulation Team Description Paper. In: RoboCup 2012 Symposium and Competitions: Team Description Papers, Mexico City, Mexico, June 2012 (2012) 

\bibitem{ggsoccer}
Kurach, K., Raichuk, A., Sta\'nczyk, P., Zajac, M., Bachem, O., Espeholt, L., Riquelme, C., Vincent, D., Michalski, M., Bousquet, O. and Gelly, S., 2020, April. Google research football: A novel reinforcement learning environment. In Proceedings of the AAAI Conference on Artificial Intelligence (Vol. 34, No. 04, pp. 4501-4510).


\bibitem{hfo}
Hausknecht, M., Mupparaju, P., Subramanian, S., Kalyanakrishnan, S. and Stone, P., 2016, May. Half field offense: An environment for multiagent learning and ad hoc teamwork. In AAMAS Adaptive Learning Agents (ALA) Workshop (Vol. 3). sn.

\bibitem{gym}
Brockman, G., Cheung, V., Pettersson, L., Schneider, J., Schulman, J., Tang, J. and Zaremba, W., 2016. Openai gym. arXiv preprint arXiv:1606.01540.

\bibitem{cython}
Behnel, S., Bradshaw, R., Citro, C., Dalcin, L., Seljebotn, D.S. and Smith, K., 2010. Cython: The best of both worlds. Computing in Science \& Engineering, 13(2), pp.31-39.

\bibitem{numpy}
Harris, C.R., Millman, K.J., Van Der Walt, S.J., Gommers, R., Virtanen, P., Cournapeau, D., Wieser, E., Taylor, J., Berg, S., Smith, N.J. and Kern, R., 2020. Array programming with NumPy. Nature, 585(7825), pp.357-362.

\end{thebibliography}
\end{document}